\documentclass[journal,10pt]{IEEEtran}
\usepackage{amsmath,amsfonts}
\usepackage{algorithm}
\usepackage{array}
\usepackage{pifont}
\usepackage[caption=false,font=normalsize,labelfont=sf,textfont=sf]{subfig}
\usepackage{textcomp}
\usepackage{stfloats}
\usepackage{verbatim}
\usepackage{algorithm}
\usepackage{algpseudocode}
\usepackage{graphicx}
\usepackage{cite}
\usepackage{url}
\usepackage{color}
\usepackage{booktabs}
\usepackage{multirow}

\usepackage{caption}
\usepackage{picins}
\usepackage{multirow}
\usepackage{rotating} 
\usepackage{booktabs} 
\usepackage{cuted}
\usepackage{float}
\usepackage{xcolor}

\usepackage[colorlinks,linkcolor=blue]{hyperref}
\usepackage{caption}

\title{Set Pivot Learning: Redefining Generalized Segmentation with Vision Foundation Models}

\author{Xinhui Li, Xinyu He, Qiming Hu, and Xiaojie Guo~\IEEEmembership{Senior Member,~IEEE}
    
\thanks{Xinhui Li (lixinhui@tju.edu.cn), Xinyu He (xy\_he68@tju.edu.cn), Qiming Hu (huqiming@tju.edu.cn) and Xiaojie Guo (Corresponding author, xj.max.guo@gmail.com) are with the College of Intelligence and Computing, Tianjin University, Tianjin 300350, China.}
}

\begin{document}

\markboth{Journal of \LaTeX\ Class Files,~Vol.~14, No.~8, August~2021}%
{Shell \MakeLowercase{\textit{et al.}}: A Sample Article Using IEEEtran.cls for IEEE Journals}

\IEEEpubid{0000--0000/00\$00.00~\copyright~2021 IEEE}

\maketitle

\vskip -30pt 

\begin{strip}  
  \centering
  \vspace{-65pt}
\includegraphics[width=\textwidth]{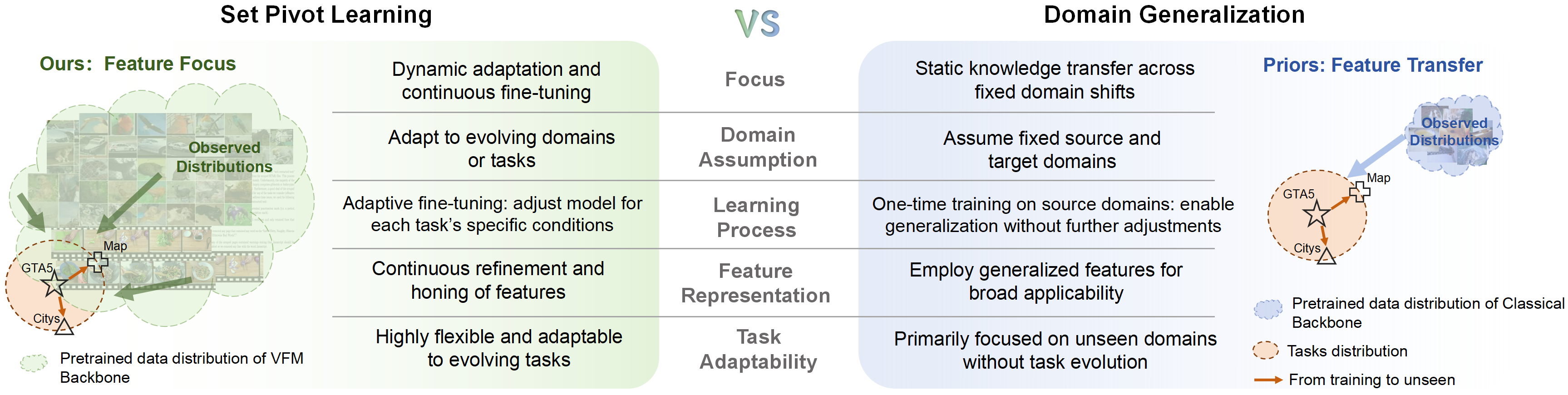}
  \captionof{figure}{Comparison between the proposed \textbf{Set Pivot Learning} and traditional Domain Generalization across key aspects.}
  \label{relatedwork}
\end{strip}

\begin{abstract}
In this paper, we introduce, for the first time, the concept of Set Pivot Learning, a paradigm shift that redefines domain generalization (DG) based on Vision Foundation Models (VFMs).
Traditional DG assumes that the target domain is inaccessible during training, but the emergence of VFMs, trained on vast and diverse data, renders this assumption unclear and obsolete. 
Traditional DG assumes that the target domain is inaccessible during training, but the emergence of VFMs, which are trained on vast and diverse datasets, renders this assumption unclear and obsolete.
To address this challenge, we propose Set Pivot Learning (SPL), a new definition of domain migration task based on VFMs, which is more suitable for current research and application requirements. Unlike conventional DG methods, SPL prioritizes adaptive refinement over rigid domain transfer, ensuring continuous alignment with evolving real-world conditions. Specifically, SPL features two key attributes: (i) Dynamic adaptation, transitioning from static domain alignment to flexible, task-driven feature optimization, enabling models to evolve with downstream scenarios; (ii) VFM-centric tuning, leveraging pretrained knowledge as a pivot to hone task-specific representations while preserving cross-domain robustness. Building on SPL, we propose a Dynamic Prompt Fine-Tuning method, which combines a Dynamic Class-aware Prompter with a Prompt-guided Feature Focuser, to elevate VFM performance in targeted scenarios. Extensive experiments on benchmark datasets show the effectiveness of our method, highlighting its superiority over state-of-the-art methods, particularly in generalized segmentation.

\end{abstract}

\begin{IEEEkeywords}
Set Pivot Learning, Vision Foundation Models, Domain Generalized Semantic Segmentation
\end{IEEEkeywords}

\section{Introduction}
\label{sec:intro}
\IEEEPARstart{R}{ecent} advances in vision foundation models (VFMs) have exhibited exceptional performance across a wide spectrum of computer vision tasks, demonstrating their superior robustness and generalization capabilities. 
Benefiting from large-scale and multimodal pretraining, VFMs have established new benchmarks across various tasks, such as image classification~\cite{naeem2023i2mvformer}, object detection~\cite{li2022grounded,zhong2022regionclip}, and semantic segmentation~\cite{wei2024stronger,Benigmim_2024_CVPR}. Building on these successes, recent research has begun to explore VFMs in more complex and unpredictable scenarios. Among these, generalized segmentation stands out, as it challenges models with real-world variability, including unknown weather conditions, dynamic environments, and a diverse array of street objects.

\begin{figure*}[t]
\centering
\includegraphics[width=1.0\textwidth]{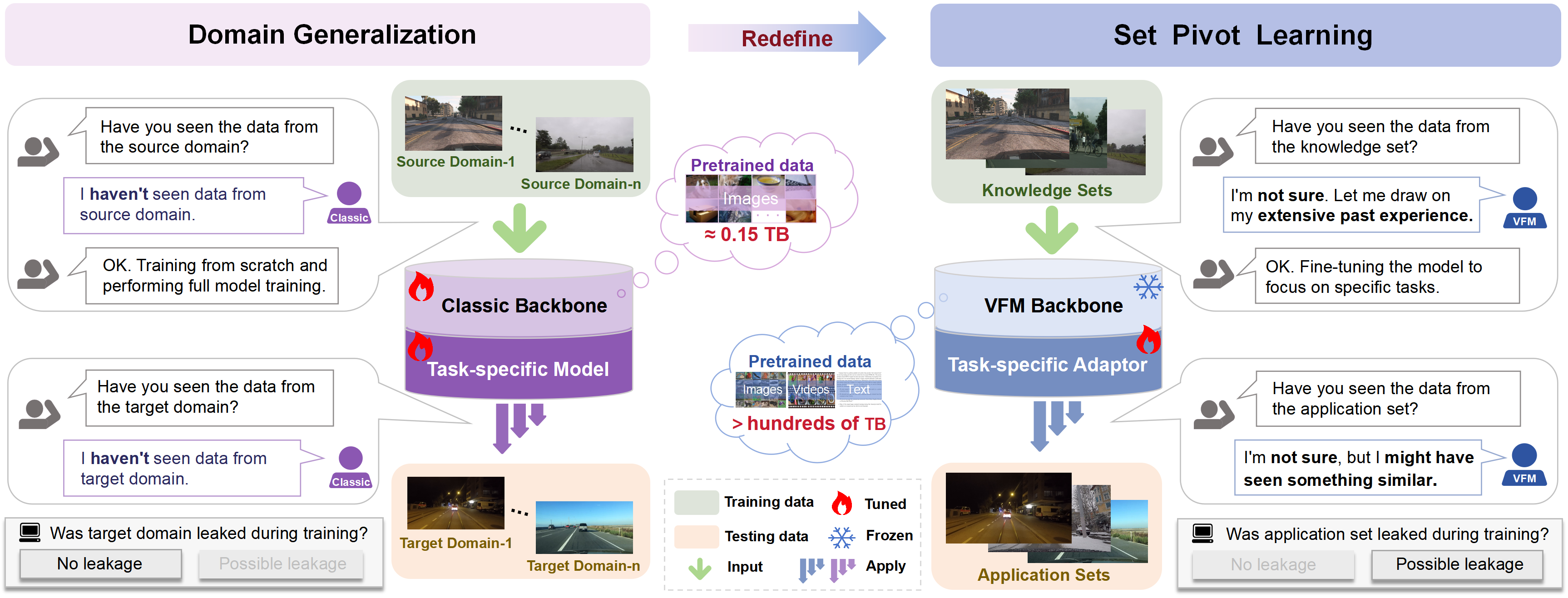} 
\caption{We introduce the concept of \textbf{Set Pivot Learning}, with the goal of leveraging the vision foundation model (VFM) as the backbone to improve the robustness and performance of the model in downstream tasks.}
\label{introduction}
\end{figure*}

Domain generalization (DG) has long been proposed to address these challenges by defining tasks with strict domain boundaries, training models on source domains and evaluating them on entirely unseen target domains.
Traditional DG methods~\cite{choi2021robustnet,huang2023style,zhao2022style,yang2023generalized}, often built upon classical network architectures, enhance generalization through data augmentation or intrinsic feature learning to mitigate distribution shifts in unseen data. However, with the emergence of VFMs, the rigid domain partitioning central to DG has grown increasingly ambiguous. As illustrated in Fig.~\ref{introduction}, VFMs undergo pre-training on vast and diverse data, making it likely that target domains are indirectly encountered or even partially supervised. This possibility raises concerns about information leakage and challenges the foundational  assumptions of DG frameworks.

\IEEEpubidadjcol

Simultaneously, the range of observed distributions in VFM-based models introduces unique challenges for generalization tasks, distinct from those faced by traditional DG methods. As shown in Fig.~\ref{relatedwork}, the primary challenges for downstream tasks have shifted from feature transfer to feature refinement and refocus. While VFMs are pretrained on large-scale, diverse datasets with strong generalization capabilities, they are not optimized for specific tasks. 
Striking a balance between the generalization strengths of Vision Foundation Models and their task-specific adaptability, such as addressing distributional discrepancies, designing effective model fine-tuning strategies, and efficiently adapting to evolving tasks, remains an open challenge.
Based on these insights, we revisit the domain generalization task and raise two key questions: (1) \textit{How should the traditional DG problem be redefined when utilizing VFM models, given the potential risk of target domain data leakage?} and (2) \textit{In the face of feature focus challenge, how can we effectively harness the generalization capabilities of VFMs for downstream tasks?}

To answer the above questions, this paper redefines the traditional DG paradigm by introducing the concept of~\textbf{Set Pivot Learning} (SPL). SPL aims to enhance model robustness by leveraging the inherent generalization capabilities of VFMs, ultimately improving performance on downstream tasks. 
Our proposed SPL is characterized by two primary features. Firstly, it employs dynamic adaptation, transitioning from static domain alignment to flexible, task-driven feature optimization, thereby enabling the model to evolve with downstream scenarios. Secondly, SPL utilizes vision foundation models as pivots, leveraging pre-trained knowledge to enhance task-specific representations while preserving cross-domain robustness.
Unlike traditional DG that imposes strict domain definitions, SPL prioritizes practical application, moving away from strict domain boundaries.
In SPL, accessible training data, such as synthetic or public datasets, are collectively referred to as the \textit{Knowledge Set}, while the intended application data forms the \textit{Application Set} (as shown in Fig.~\ref{introduction}).
Furthermore, we investigate challenges emerging within this framework. 
As shown in Table~\ref{tab:syn-real}, our experiments reveal that even without fine-tuning, frozen VFMs can significantly outperform previous generalization methods.
This finding not only validates the inherent generalization capabilities of VFMs, but also shifts the core challenge from traditional model improvement or data augmentation toward strategically magnifying VFMs' advantages for downstream tasks.

To tackle the mentioned challenge, we propose a novel dynamic prompt fine-tuning method that generates prompts from the knowledge set to effectively transfer VFMs' generalization capabilities to task-specific optimization. Specifically, we design a Dynamic Class-aware Prompter (DCP) to flexibly extract scene-specific prompts from images through category filtering and hierarchical clustering. In addition, we propose a Prompt-guided Feature Focuser (PFF) integrated within the VFM layers, which utilizes similarity prompts, class prompts and a multimodal attention mechanism to fine-tune the VFM. Such a selective fine-tuning strategy enhances model robustness and performance for generalized segmentation.
The main contributions of this work can be summarized as follows:
\begin{itemize}
    \item We redefine the domain generalization task and introduce \textbf{Set Pivot Learning}, which aims to leverage the pre-trained knowledge of VFMs as pivot to dynamically adapt and enhance the robustness of the downstream tasks.
    \item We propose a dynamic prompt fine-tuning method with a dynamic class-aware prompter and prompt-guided feature focuser to effectively adapt the generalization strengths of VFMs to specific tasks.
    \item Extensive experiments are conducted to demonstrate the rational of our novel framework, and reveal its superior performance in generalized street scene segmentation.
\end{itemize}

\section{Related Work}
\subsection{Domain Generalization}
\label{sec:DA}
To address the challenge of performance degradation when models encounter data from unseen domains, domain generalization (DG) has emerged as a key strategy. DG methods~\cite{su2022consistency, li2022cross,10144687} are fundamentally designed to enhance the generalization of models under diverse or unknown distribution conditions.
The field of DG can be broadly categorized into two main branches: multiple-source DG and single-source DG.
In multiple-source DG~\cite{Matsuura2020DomainGU,Wang2021GeneralizingTU,Liu2021FedDGFD,Li2021SemanticSW}, the model is trained on multiple source domains and then encourage the network to learn domain-invariant feature representations over diverse distributions.
Feature alignment-based approaches~\cite{Peng2019MomentMF,Zhu2021SelfsupervisedUD} have gained popularity in multiple-source DG, where distribution constraints are enforced through techniques such as normalization and whitening~\cite{Nam2018BatchInstanceNF,Jia2019FrustratinglyEP}.
Various learning strategies have been widely employed to enhance the generalization ability of models.
Meta-learning approaches~\cite{Li2019EpisodicTF,Shu2021OpenDG,Kim2022PinTM}, such as episodic training and open-set domain generalization, have shown promising results in learning adaptable models.  Adversarial feature learning~\cite{Li2018DomainGW} techniques have also been utilized to learn domain-invariant representations, and metric learning\cite{Motiian2017UnifiedDS,dou2019domain} methods have played a role in enhancing the discriminative power of the models across domains. 
Although these methods address the unavailability of the target domain by utilizing multiple source domains, in real-world scenarios, collecting data from multiple domains can be challenging. Therefore, there is a growing interest in studying domain generalization with a single-source approach.

Single-source DG methods have been developed to address the challenges of using only a single source domain during training.
Motivated by the multiple-source DG techniques, some single-source DG methods~\cite{learningli,Xu2022DIRLDR,wu2022siamdoge,10539285} strive to extend and augment the single source data to improve generalization performance.
\cite{Volpi2018GeneralizingTU} designed an iterative scheme that enables adaptive data augmentation by appending adversarial examples at each iteration. 
\cite{Qiao2020LearningTL} proposed adversarial domain augmentation to guide the model learning generalized knowledge.
\cite{Wang2021LearningTD} improved the generalization of the network by learning to generate diverse images and designing mutual information optimization strategy.
Additionally, single-source DG methods have explored techniques such as style transfer~\cite{Prakash2019StructuredDR,Peng2021GlobalAL} and domain randomization~\cite{Yue2019DomainRA,Huang2021FSDRFS} to augment the single source data and extend the input distribution. 
Moreover, feature normalization~\cite{Pan2018TwoAO,Seo2020LearningTO} has been investigated as an effective strategy to exploit potential feature representations. By normalizing the features, models can mitigate the over-fitting to the training data and improve generalization performance. 
Meanwhile, feature disentanglement is studied to separate the feature representations into domain-variant and domain-invariant components~\cite{choi2021robustnet,peng2022semantic} to improve generalization performance.
On the other hand, singlesource DG focuses on training the model using a single source domain, with the aim of generalizing to the target domain, often under the assumption of minimal domain shift. Both approaches aim to enhance the model’s robustness to domain variability, but they differ in the way they utilize source domain data to improve generalization performance.

Most DG methods~\cite{li2022cross, chen2022learning, jiang2023domain, Fahes_2024_CVPR} are built on classical network backbones, such as ResNet~\cite{he2016deep}, VGGNet~\cite{simonyan2014very}, and ShuffleNetV2~\cite{ma2018shufflenet}, typically pre-trained on datasets like the ImageNet~\cite{deng2009imagenet}.
These methods assume the network has not been exposed to real-world domains like street scenes, a foundational premise for research on model robustness. However, with the rise of more advanced backbones, such as Vision Foundation Models (VFMs)~\cite{vidit2023clip, he2022mae, kirillov2023sam, fang2023eva01, fang2024eva02, oquab2023dinov2}, trained on vast and diverse datasets, the traditional assumption of domain invisibility in DG is becoming increasingly uncertain.
Consequently, this paper redefines the problem of generalized segmentation within the VFM framework, addressing the unique challenges posed by these more generalized and pre-trained models.

\subsection{VFMs in Generalized Segmentation}
Benefiting from training on large-scale and diverse datasets, Vision Foundation Models (VFMs), such as CLIP~\cite{radford2021clip}, SAM~\cite{kirillov2023sam}, and DINOV2~\cite{oquab2023dinov2}, exhibit strong performance across a wide range of vision tasks~\cite{vidit2023clip, lai2024lisa, ma2024segment, chen2024anydoor, chen2024end}.
Recent studies~\cite{li2024learning, wei2024stronger, benigmim2024collaborating} reveal that using VFMs as backbones for downstream tasks, including object detection and semantic segmentation, yields significant performance improvements even without fine-tuning on specific domain data. For example, the work~\cite{wei2024stronger} indicates that VFMs can rival previous methods, even without fine-tuning on specific street scene data, as shown in Fig.~\ref{relatedwork}.
Furthermore, prompt-based fine-tuning methods~\cite{fahes2024simple, kim2024eclipse, Fahes_2023_ICCV} have been shown to align the general knowledge from VFMs with task-specific needs, enhancing their generalization across different tasks. Other strategies, such as domain adaptation~\cite{wei2024stronger}, contrastive learning~\cite{ge2023domain}, and multi-modal learning~\cite{li2024split, benigmim2024collaborating}, have also been explored to improve performance in generalization tasks.

However, despite the promising results, applying VFMs to generalized segmentation tasks remains underexplored. current methods~\cite{wei2024stronger, gong2024coda, vidit2023clip, yang2024unified, fahes2024simple,pak2024textual} primarily emphasize fine-tuning or leveraging VFMs often neglecting  traditional domain generalization definition. VFMs, however, may already possess implicit knowledge about the target domain due to their pre-training on diverse datasets. This suggests the need for a reevaluation of domain generalization in the VFM context, and focusing on how to leverage the pre-existing knowledge to improve generalized segmentation. 
Therefore, we propose the concept of set pivot learning in this paper, which leverages the the general knowledge from VFMs as pivot and dynamically adjust pivot to downstream tasks.

\subsection{Prompt Learning}
Inspired by the success of prompt learning in natural language processing, prompt learning~\cite{ge2023domain, khattak2023maple, xin2024mmap, bahng2022exploring} has emerged as a promising approach in computer vision tasks. Traditional methods~\cite{li2022cross, zhao2023masked, bi2024learning} for adapting foundation models to downstream tasks often involve retraining, which is resource-intensive and time-consuming. To address these limitations, prompt tuning~\cite{jia2022visual, lester2021power, ge2023domain} has been introduced as a more efficient alternative, which enhances the generalization performance in street scene recognition tasks by providing corresponding text prompts and then fine-tuning the network. Recent works~\cite{Fahes_2023_ICCV, yang2024unified, vidit2023clip, fahes2024simple, gong2024coda, kim2024eclipse, li2024learning} have proven the substantial performance gains achievable through prompt learning strategies. Specifically, the success of pre-trained vision-language models like CLIP~\cite{radford2021clip} has spurred interest in designing suitable prompts for street scene recognition. For instance, schemes in~\cite{Fahes_2023_ICCV, yang2024unified, fahes2024simple} utilize manually crafted prompts to drive zero-shot domain adaptation, enabling effective model adaptation to new domains.

Further, some methods~\cite{li2024learning, bai2024prompt, xin2024mmap} leverage learnable prompts to guide and enhance domain feature representations. These learnable approaches not only capture domain-specific nuances but also allow for dynamic adjustment during fine-tuning, further improving adaptability. Despite their effectiveness, these techniques often demand considerable expertise and effort in prompt design. In particular, manually creating and fine-tuning prompts can be an arduous and time-consuming process, especially when dealing with complex or highly variable domain-specific data, such as street scene recognition. To streamline this process, this paper introduces a dynamic prompt fine-tuning strategy that automatically generates task-specific prompts tailored to the street scene context. By leveraging this adaptive mechanism, our method significantly improves the robustness and generalization of the model in practical applications, reducing the burden of manual prompt design and fine-tuning.

\section{Set Pivot Learning}
We are the first to investigate the problem settings and differences between classical backbones and vision foundation model backbones in the domain generalization task.
Building on the distinctive attributes of VFMs, pre-trained on massive datasets, we first propose the concept of \textbf{Set Pivot Learning}. 
The following subsections will elaborate on the problem statement and the associated notations.

\begin{figure*}[t]
\centering
\includegraphics[width=1.0\textwidth]{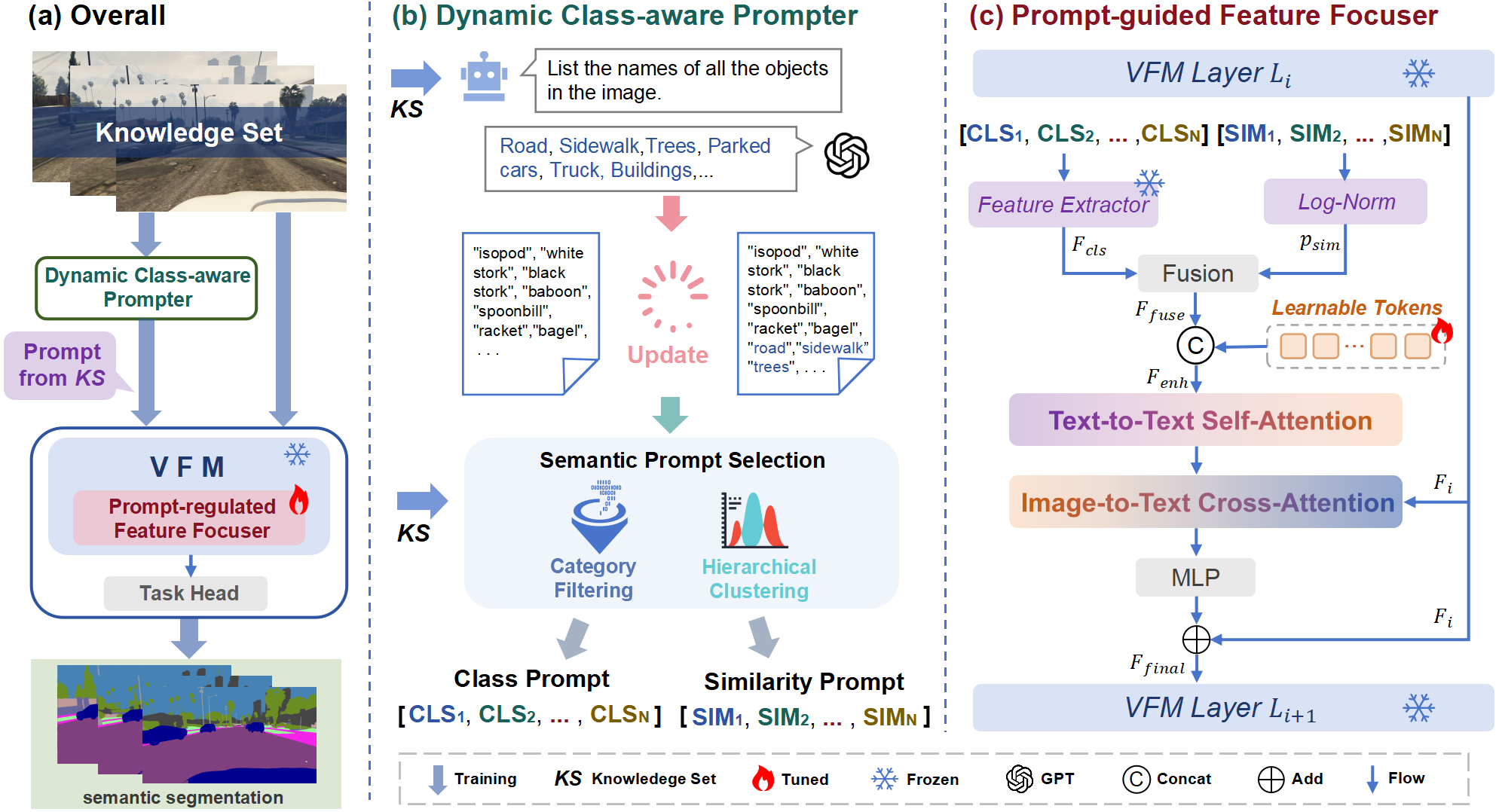} 
\caption{An overview of our method. \textbf{(a)} Network architecture. \textbf{(b)} Dynamic Class-aware Prompter. \textbf{(c)} Prompt-guided Feature Focuser.}
\label{network}
\end{figure*}

\subsection{Problem Statement}

Set Pivot Learning aims to tackle the challenge of strengthening the robustness and generalization of models across diverse domains. As discussed in Sec.~\ref{sec:DA} and summarized in Fig.~\ref{relatedwork}, traditional Domain Generalization (DG) methods rely on a clear distinction between source and target domains. However, in the context of VFM, this distinction becomes vague due to the extensive and diverse nature of training data. Therefore, existing methods~\cite{wei2024stronger,pak2024textual} still use DG to name these generalization tasks are not rigorous.
To contextualize, we examine the generalized semantic segmentation task. In traditional setups, DG involves training on a source domain (\textit{e.g.}, GTA5~\cite{Richter2016PlayingFD}) and evaluating on unseen target domains (\textit{e.g.}, Cityscapes~\cite{Cordts2016TheCD} and Mapillary~\cite{Neuhold2017TheMV}). Most DG methods~\cite{Peng2021GlobalAL,peng2022semantic,wu2022siamdoge,li2023atf} typically employ classical backbones such as ResNet-50 or ResNet-101. They are typically initialized with parameters pre-trained on the ImageNet dataset~\cite{krizhevsky2012imagenet}, which is a large-scale dataset of natural images with 14,197,122 entries. It facilitates the initial training of networks for semantic segmentation task while strictly adhering to the DG task's setup of target domain invisibility. However, with the advent of VFMs, pretraining spans vast and varied datasets, challenging the foundational assumption of target domain invisibility in traditional DG.
For instance, several methods~\cite{wei2024stronger, benigmim2024collaborating, li2024learning} employ backbones such as CLIP~\cite{radford2021clip}, SAM~\cite{kirillov2023sam}, or DINOv2~\cite{oquab2023dinov2}. These VFMs are trained on datasets of hundreds of terabytes or even larger, encompassing not only images but also a wide range of modalities, including videos and texts.

Notably, as shown in Fig.~\ref{relatedwork}, even frozen VFMs generally outperform the proposed methods that use classical backbones in the semantic segmentation DG task.
These performances prompt us to consider two important questions: \textit{Are these superior generalization performances in downstream tasks attributable to the vast amount of data involved in VFM training? Has the target domain been inadvertently exposed within this extensive data?}
However, these questions are difficult to answer definitively.
Given that VFM backbones are pre-trained on hundreds of terabytes of publicly or privately data, it becomes challenging to ensure that target domain data have not been included in these extensive datasets. 
This uncertainty challenges the original premise of domain invisibility, rendering the traditional DG definition less applicable to VFM-based methods.
Consequently, the original premise of target domain invisibility is difficult to uphold and the traditional definition DG is no longer well-suited to these VFM-based methods.

In light of these considerations, we present a new concept, termed \textbf{Set Pivot Learning} (SPL), based on VFM backbones. Given that the vast amounts of data have blurred the boundaries between domain distributions, SPL shifts the focus away from strictly enforcing domain visibility or invisibility. Instead, SPL is driven by the practical objective of solving model robustness problems in the real-world, with the primary goal of enhancing the model's generalization capabilities. In SPL, the model leverages the knowledge acquired from large pre-trained VFMs as a pivot, and then dynamically adapts to downstream tasks.
In the setup of SPL, we categorize data into two \textit{sets}. The data available for training is defined as ``Knowledge Sets" and the intended deployment domains as ``Application Sets". In the following, we will provide a detailed explanation of these concepts.

\subsection{Knowledge Set \& Application Set}
Knowledge Set (KS) is defined as the collection of labelled data available for training in set pivot learning. Specifically, KS is no longer limited to a specific domain where data shares the same distribution. In practical applications, all available training data are considered part of the KS. For instance, in street scene recognition tasks, the available datasets typically include the real-world dataset, synthetic dataset, and potentially a growing number of private or unpublished street scene datasets. Therefore, we introduce a broader concept, \textbf{sets}, which no longer adheres to strict data distributions or dataset partitions, but instead prioritizes solving real-world tasks. This also aligns with the current trend of leveraging diverse datasets for model training in practical applications.

Application Set (AS) refers to the data intended to deploy the model, that is, the data used to evaluate the model's performance in the experiment. Unlike the KS composed of labeled data for training, AS are characterized by greater diversity and unpredictability. However, in the context of extensive pre-training on large-scale datasets by large models, the data in the AS is no longer entirely unknown. Consequently, the strict invisibility of the target domain, a traditional assumption in DG, is increasingly difficult to uphold. Therefore, from a practical application perspective, we are no longer constrained by data invisibility.

\section{Methodology}
Building on set pivot learning, we propose a dynamic prompt fine-tuning method that effectively adapts the generalization of VFMs to specific tasks. In this process, our approach mainly focuses on answering the two questions asked in Sec.~\ref{sec:intro}.: (1) How to effectively generate task-specific prompts that accurately capture the unique class of the input image, and (2) Based on these prompts, how to efficiently transfer and adapt the VFM’s generalization capabilities to the new scenarios. In the following, we present our solutions to these challenges, detailing the design and implementation.

\begin{algorithm}[t]
\caption{Semantic Prompt Selection for DCP}
\label{alg:filter_clustering}
\textbf{Input:} CLIP model $\phi$, input image $I$, class library $\mathcal{C}$, filter threshold range $[\tau_{f}^{\min}, \tau_{f}^{\max}]$, cluster threshold range $[\tau_{c}^{\min}, \tau_{c}^{\max}]$, filter increment $\Delta \tau_f$, cluster increment $\Delta \tau_c$, initial label count $L = |\mathcal{C}|$, maximum output classes $N$.\\
\textbf{Output:} Class prompts $\mathcal{P}_{\text{cls}}$ and their associated similarity prompts $\mathcal{P}_{\text{sim}}$.

\begin{algorithmic}[1]

    \State Compute similarity scores $S = \{s_l \mid l \in \mathcal{C}\}$ between $I$ and each class in $\mathcal{C}$ using $\phi$.

    \State Initialize $\tau_f \gets \tau_{f}^{\min}$, $\tau_c \gets \tau_{c}^{\min}$.

    \While{$L > N$ \textbf{and} $\tau_f \leq \tau_{f}^{\max}$ \textbf{and} $\tau_c \leq \tau_{c}^{\max}$}

        \State  \texttt{$\triangle$ Category Filtering}
        \State Initialize $\mathcal{P}_{\text{cls}}\gets [\hspace{0.1cm}]$, $\mathcal{P}_{\text{sim}}\gets [\hspace{0.1cm}]$.
        \For{each $l \in L$}
            \If{$s_l > \tau_f$}
                \State Add $l$ to $\mathcal{P}_{\text{cls}}$ , $s_l$ to $\mathcal{P}_{\text{sim}}$
            \EndIf
        \EndFor

        \State  \texttt{$\triangle$ Hierarchical Clustering}
        \State Perform hierarchical clustering on $\mathcal{P}_{\text{cls}}$ using features from $\phi$ with the stopping criterion $\tau_c$.
        \State Update $\mathcal{P}_{\text{cls}}$ and $\mathcal{P}_{\text{sim}}$ by selecting the most central class in each cluster.
        \State Update $L \gets |\mathcal{P}_{\text{cls}}|$

        \State Increment $\tau_f \gets \tau_f + \Delta \tau_f$, $\tau_c \gets \tau_c + \Delta \tau_c$.
    \EndWhile
    \State \textbf{Return} $\mathcal{P}_{\text{cls}}$, $\mathcal{P}_{\text{sim}}$.
\end{algorithmic}
\end{algorithm}

\subsection{Network Architecture}
The overall architecture of the proposed method is illustrated in Fig.~\ref{network}~(a). Let \( \mathcal{S}^K \) and \( \mathcal{S}^A \) represent the knowledge set and application set, respectively. Specifically, we define the KS as \( \mathcal{S}^K = \{ (x_i, y_i) \}_{i=1}^{N} \), where each data sample \( (x_i, y_i) \) consists of an input image \( x_i \) and its corresponding label \( y_i \), with \( N \) denoting the total number of samples in \( \mathcal{S}^K \).
Similarly, the AS is defined as \( \mathcal{S}^A = \{ (x_j, y_j) \}_{j=1}^{M} \), where \( M \) is the number of samples in the application set. The goal of SPL is to train a model \( \varphi_\theta(\cdot) \) using only the \( \mathcal{S}^K \), ensuring that it generalizes effectively to the \( \mathcal{S}^A \). 
The model's effectiveness is validated on unseen \( \mathcal{S}^A \) after being trained on \( \mathcal{S}^K \). Firstly, we propose a Dynamic Class-aware Prompter (DCP) module, which is designed to dynamically extract category-specific prompts from the \( \mathcal{S}^K \). Additionally, we introduce a Prompt-guided Feature Focuser (PFF), integrated within the VFM layers. 
The PFF leverages the class prompt and similarity prompt to fine-tune the VFM, enhancing its applicability to the segmentation task. Specifically, given an input image $I$, the DCP module dynamically generates prompts based on the \( \mathcal{S}^K \), including both class prompts and their associated similarity prompts. These generated prompts are subsequently processed by the trainable PFF module, while the VFM remains in a frozen state. By fine-tuning the PFF, the network is guided to focus more effectively on learning feature representations relevant to the downstream task, enabling the model to better adapt to real-world scenarios. 
Finally, the task head is responsible for generating the segmentation predictions. 
In the testing, replacing the input with the application set can adapt the model to diverse use cases.

\begin{figure*}[t]
\centering
\includegraphics[width=1.0\textwidth]{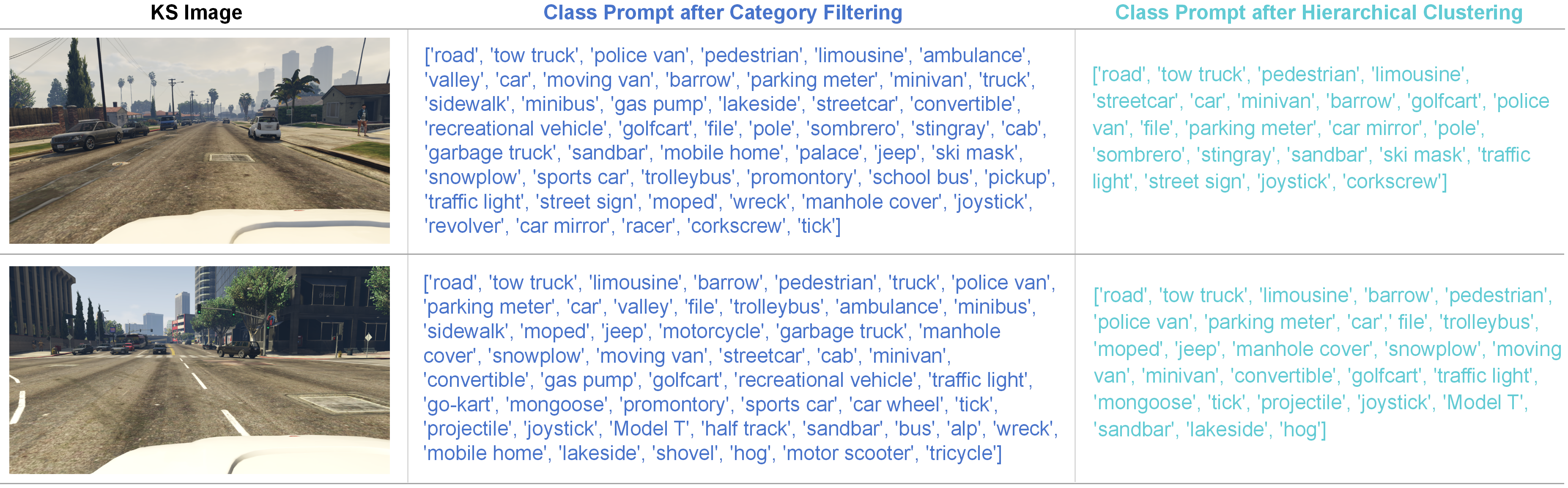} 
\caption{The class prompt for KS images (from the GTA5 dataset) after category filtering and hierarchical clustering, respectively. The maximum output classes $N$ of DCP set to 30.}
\label{prompt}
\end{figure*}

\subsection{Dynamic Class-aware Prompter}
As a large-scale model, the vision foundation model is pre-trained on a huge number of different classes or categories~\cite{radford2021clip}. However, focusing on specific downstream tasks and their associated classes is critical to achieving optimal performance. In the task of street scene recognition, narrowing the focus to relevant categories while ignoring those unlikely to appear in the street becomes particularly important. To address this, we propose a Dynamic Class-Aware Prompter (DCP) that adaptively generates prompts based on the input image in an unsupervised manner. Specifically, DCP dynamically identifies related classes, generates class-specific prompts along with their corresponding similarity prompts, and provides guidance for subsequent network fine-tuning. 
The core concept of the DCP is to adaptively select and generate class prompts for $I$ from a category library in an unsupervised manner. The architecture of the DCP is depicted in Fig.~\ref{network}~(b).

\noindent\textbf{Initial Category Library.}
To construct a broad category selection space, we first use the 1000 classes of ImageNet~\cite{krizhevsky2012imagenet} as the initial category library \(\mathcal{C}_{\text{init}}\). 
However, considering that the categories in the knowledge set may extend beyond the scope of \(\mathcal{C}_{\text{init}}\), we further supplement \(\mathcal{C}_{\text{init}}\) through GPT-based queries, as illustrated in Fig.~\ref{network}~(b). 
In the experiments, BLIP~\cite{li2022blip} can also be employed as an alternative to GPT for visual question answering to enrich the category library. It should be emphasized that the sole purpose of invoking GPT is to supplement library categories during the knowledge preparation phase, and no subsequent GPT queries are required in the training and testing process. 
Specifically, we implement this by selecting a limited number of images from the knowledge set for GPT-based interrogation prior to model training, thereby completing the library supplement process. 
The computational and financial costs incurred by this process are almost negligible, as adding the category library is a one-time preprocessing step performed prior to training.
After this step, we obtain the final category library \(\mathcal{C}\) used for retrieval. It is worth noting that the category library \(\mathcal{C}\) does not need to be updated during training and testing. The purpose of our initialization is to come up with a flexible way to deal with situations where there may be insufficient categories in the \(\mathcal{C}\) compared to the actual categories in KS.

Subsequently, we design a semantic prompt selection strategy, which selects and generates prompts within the \(\mathcal{C}\) space.
The proposed strategy consists of two key components: category filtering and hierarchical clustering~\cite{müllner2011modernhierarchicalagglomerativeclustering}. 
These two components work iteratively to refine the prompt set, aligning it with task-specific categories.

\noindent\textbf{Category Filtering.} 
This step is responsible for selecting the most relevant categories from a broad category library. Specifically, the similarity scores between the input KS and all class prompts in the $\mathcal{C}$ are firstly computed using the CLIP model. These similarity scores are then passed through a softmax operation to obtain the probabilities (SIM) for each class (CLS). By setting probability threshold $\tau_{f}$, class prompt with probabilities below the threshold are filtered out, while the remaining class prompt and their corresponding probabilities are retained. The detailed process is provided in Algorithm~\ref{alg:filter_clustering}.

\noindent\textbf{Hierarchical Clustering.} After the filtering process, we often encounter numerous class prompts with high semantic similarity, such as ``minibus" and ``minivan", ``convertible" and ``sports car". This leads to the problem of prompt redundancy. To address this issue, we adopt hierarchical clustering~\cite{müllner2011modernhierarchicalagglomerativeclustering} to further reduce the number of class prompts. Specifically, we compute text features using the CLIP model and iteratively adjust the distance threshold for clustering by setting a threshold range and step size. As described in Algorithm~\ref{alg:filter_clustering} and shown in the example in Fig~\ref{prompt}, we extract the representative class with the highest intracluster similarity to the cluster center, thereby minimizing redundancy and ultimately obtaining the optimal clustering result.

These process result in a filtered set of class prompts \(\mathcal{P}_{\text{cls}}\) and their corresponding similarity prompts \(\mathcal{P}_{\text{sim}}\).
Intuitively, our DCP module can be viewed as a general and efficient filtering mechanism. 
Starting with category mining from images, it extracts the categories that the VFM should focus on for downstream tasks in an unsupervised manner. 
This dynamic prompt generation enables the subsequent fine-tuning module to be guided by more relevant and targeted categories for each image, thereby facilitating more effective task adaptation and improved performance.

\subsection{Prompt-guided Feature Focuser}
To effectively leverage dynamically generated prompts from the DCP and facilitate the transfer of VFM's generalized capabilities to downstream tasks, we propose a Prompt-guided Feature Focuser (PFF). The core innovation of the PFF module lies in its use of similarity and class prompts to guide the network in fine-tuning feature representations for specific scenarios, thereby adaptively focusing on the relevant features for downstream tasks. Specifically, the PFF incorporates a prompt fusion mechanism, text-to-text self-attention and image-to-text cross-attention modules, as depicted in Fig.~\ref{network}~(c).

\textbf{Similarity-Based Prompt Fusion.}
Based on the obtained class prompts \(\mathcal{P}_{\text{cls}}\) and their corresponding similarity prompts \(\mathcal{P}_{\text{sim}}\), we design a feature fusion mechanism based on similarity embeddings. 
Initially, the \(\mathcal{P}_{\text{cls}}\) is processed by the
CLIP text encoder~\cite{radford2021clip} into the feature embedding \(F_{\text{cls}}\).
We then apply logarithmic normalization to \(\mathcal{P}_{\text{sim}}\) to ensure numerical stability, resulting in \(p_{\text{sim}}\).
The \(p_{\text{sim}}\) is then fused with \(F_{\text{cls}}\), leading to $F_{\text{fuse}}$ as:
\begin{equation}
    F_{\text{fuse}} = \text{MLP} \left( F_{\text{cls}} \circ \texttt{expand}(p_{\text{sim}}, \, \text{dim} = d)\right),
\end{equation}
where the \texttt{expand} operation replicates \( p_{\text{sim}} \) to the same dimension \( d \) of $F_{\text{cls}}$ to match the required shape for subsequent operations. Here,  MLP represents Multi-Layer Perceptron, and $\circ$ denotes the element-wise product.

\begin{table*}[htbp]
\centering
\renewcommand\arraystretch{1.3} 
\setlength\tabcolsep{8pt} 
\resizebox{\textwidth}{!}{
    \begin{tabular}{c|cccc|cccc|cccc}
    \toprule
    \multirow{2}{*}{\textbf{Methods}} & \multicolumn{4}{c|}{\textbf{Classical-based Backbone (ResNet-50)}} & \multicolumn{4}{c|}{\textbf{VFM-based Backbone (EVA02)}} & \multicolumn{4}{c}{\textbf{VFM-based Backbone (DINOV2)}} \\
    \cmidrule(lr){2-5} \cmidrule(lr){6-9} \cmidrule(lr){10-13}
     & \textbf{Cityscapes} & \textbf{BDD-100K} & \textbf{Mapillary} & \textbf{Avg.} & \textbf{Cityscapes} & \textbf{BDD-100K} & \textbf{Mapillary} & \textbf{Avg.} & \textbf{Cityscapes} & \textbf{BDD-100K} & \textbf{Mapillary} & \textbf{Avg.} \\
    \midrule
    \textbf{Backbone only} & 28.9 & 25.3 & 29.3 & 27.8 & 56.5 & 53.6 & 58.6 & 56.2  & 63.3 & 56.1 & 63.9 & 61.1\\
    \midrule
    RobustNet~\cite{choi2021robustnet} & 37.7 & 34.1 & 38.5 & 36.8 & -- & -- & -- & -- & -- & -- & -- & -- \\
    AdvStyle~\cite{zhong2022adversarial} & 39.6 & 35.5 & 37.0 & 37.4 & 51.4 & 51.6 & 56.5 & 53.2 & 61.5 & 55.1 & 63.9 & 60.1 \\
    GTR~\cite{Peng2021GlobalAL} & 37.5 & 33.8 & 34.5 & 35.3 & 52.5 & 52.8 & 57.1 & 54.1 & 60.2 & 57.7 & 62.2 & 60.0 \\
    PASTA~\cite{Chattopadhyay_2023_ICCV} & 41.2 & 38.4 & 41.6 & 40.4 & 57.8 & 52.3 & 58.5 & 56.2 & 62.1 & 57.2 & 64.5 & 61.3 \\
    VPT~\cite{jia2022visual} & -- & -- & -- & -- & 62.2 & 57.7 & 62.5 & 60.8 & 65.2 & 59.4 & 65.5 & 63.3 \\
    Rein~\cite{wei2024stronger} & -- & -- & -- & -- & 65.3 & 60.5 & 64.9 & 63.6 & 66.4 & 60.4 & 66.1 & 64.3 \\
    \textbf{Ours} & -- & -- & -- & -- & \textbf{66.7} & \textbf{61.7} & \textbf{65.8} & \textbf{64.7} & \textbf{67.9} & \textbf{61.8} & \textbf{67.6} & \textbf{65.8} \\
    \bottomrule
    \end{tabular}%
}
\caption{Comparison of mIoU ($\%$) with other DGSS methods using classical-based and VFM-based backbones. The model is trained on the synthetic dataset (GTA5) and generalized to real-world datasets (Cityscapes, BDD-100K, Mapillary).}
\label{tab:syn-real}
\end{table*}

\textbf{Cross-Modal Interaction Enhancement.}
To guide the model in focusing and learning fused features
$F_{\text{fuse}}$, we employ cross-modal interactive attention to enhance the model's capabilities. This process primarily includes text-to-text self-attention and image-to-text cross-attention.

\noindent\textit{Text-to-Text Self-Attention.}
After the fusion, we employ a set of learnable tokens \( T = \{ T_i \in \mathbb{R}^{m \times c} \mid 1 \leq i \leq N \} \), where each token sequence \( T_i \) is randomly initialized. \( m \) denotes the sequence length of \( T_i \). \( c \) is the dimensionality of \( F_i\). \( N \) denotes the number of VFM layers. And then we concat the \( T_i\) with $F_{\text{fuse}}$ to get the enhanced feature \( F_{\text{enh}} = \text{concat}(T_i, F_{\text{fuse}}) \). The fused feature representation, which combines the class prompt and the similarity prompt, is processed using a multi-head self-attention mechanism. Given $F_{\text{enh}}$, the self-attention outputs the weighted sum of values $V_{text}$ based on learned attention scores: \( \text{SelfAtt} (Q_{text}, K_{text}, V_{text}) = \text{softmax}\left( \frac{Q_{text} K_{text}^\top}{\sqrt{d_k}} \right) V_{text}\), where $Q_{text}, K_{text}, V_{text}$ are linear projections of $F_{\text{enh}}$, and $d_k$ is the dimension of each key vector. In implementation, the self-attention captures intra-modal dependencies within the fused feature set, refining the features based on intra-prompt attention scores.

\noindent\textit{Image-to-Text Cross-Attention.}
The output from the self-attention module is then fed into a cross-attention layer to incorporate image features $F_i$, which is produced by the \( i \)-th VFM layer \( L_i \). The cross-attention operates as follows: \( \text{CrossAtt} (Q_{text}, K_{img}, V_{img}) = \text{softmax}\left(\frac{Q_{text} K_{img}^\top}{\sqrt{d_k}}\right) V_{img}\)
, where $Q_{text}$ is derived from the output of the self-attention mechanism, while $K_{img}$ and $V_{img}$ originate from the image features $F_i$. This step enables the model to incorporate cross-modal interactions, adapting the fused features based on image characteristics.
After obtaining the cross-attended features, they are sent into a Multi-Layer Perceptron (MLP). The final feature $F_{\text{final}}$ can be represented as:
\begin{equation}
    F_{\text{final}} = \text{MLP}\left(\text{CrossAtt}\left(\text{SelfAtt}(F_{\text{enh}})\right)\right) + F_i.
\end{equation}

\textbf{PFF Fine-Tuning.} 
The PFF module is integrated into the VFM's feature layers\( L_i \) ($i\in\{1,2,...,N\}$), where \( N \) is the number of layers. During training, the feature layers \( L_1, L_2, \ldots, L_N \) are kept frozen, allowing for the fine-tuning of only the PFF module to facilitate the transfer of the large model to downstream tasks. 
In this study, we incorporate a task-specific head, Mask2Former~\cite{Cheng2021MaskedattentionMT}, for semantic segmentation and adopt its loss configuration throughout our experiments to mitigate potential sources of interference.
Therefore, by integrating class and similarity prompts with attention mechanisms, the model achieves robust and flexible feature modulation and focusing.

\section{Experimental Validation}
In this section, we provide a comprehensive overview of the implementation details and present the experimental results of our method. 
We evaluate and compare the performance of our work in semantic segmentation tasks under both normal and adverse conditions with recent methods. 
Meanwhile, we analyze the effectiveness of the designed modules and visualize the segmentation results.

\subsection{Visual Foundation Models}
In this study, we conduct a comprehensive evaluation of the impact of Visual Foundation Models (VFMs) in generalized segmentation. To this end, we analyze five distinct VFMs: MAE~\cite{he2022mae}, CLIP~\cite{radford2021clip}, SAM~\cite{kirillov2023sam}, EVA02~\cite{fang2023eva02,fang2023eva}, and DINOv2~\cite{oquab2023dinov2}. Specifically, CLIP is a language-image pre-training model. MAE employs a masked image modeling approach to learn effective visual representations. SAM is trained on a large-scale segmentation dataset, aiming to enhance the model's generalization across various segmentation tasks. EVA02 combines the CLIP framework with masked image pre-training. DINOv2 is a self-supervised model pre-trained on a curated image dataset. For consistency in terms of both performance and computational efficiency, we primarily use the ViT-Large architecture across all VFMs, with the exception of SAM, which relies on a ViT-Huge image encoder as described in the original work~\cite{kirillov2023sam}.

\begin{figure*}[t]
\centering
\includegraphics[width=1.0\textwidth]{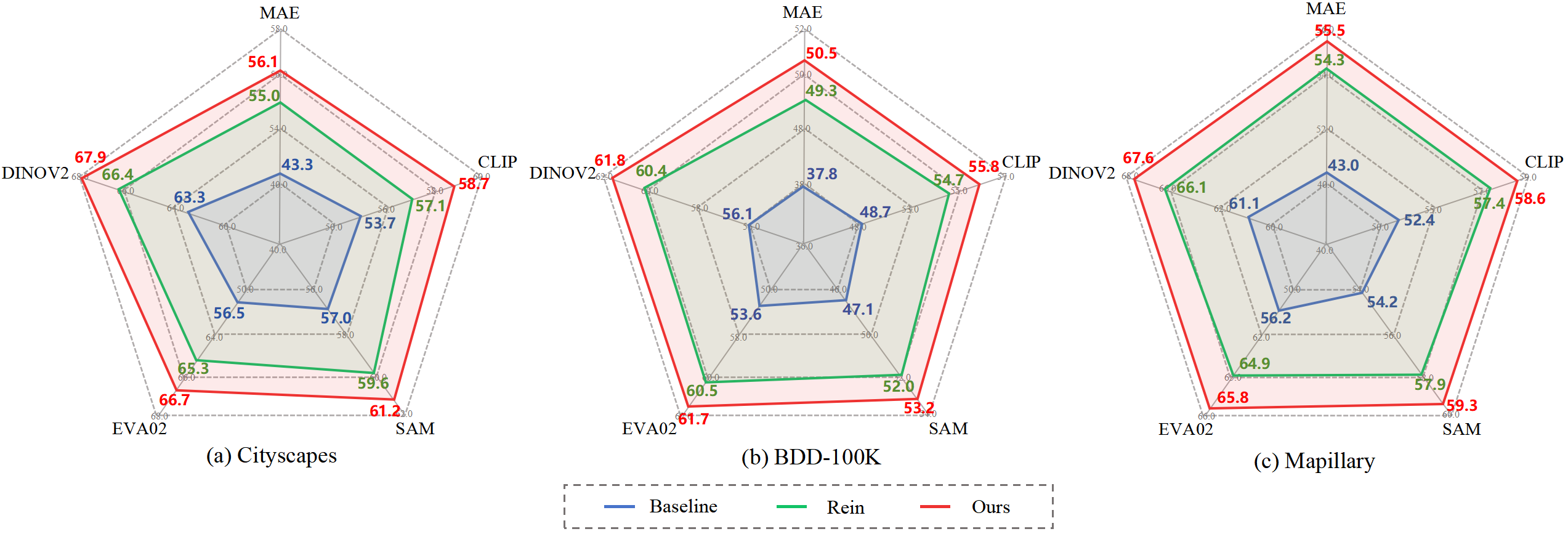} 
\caption{Performance comparison with multiple VFMs (MAE, CLIP, SAM, EVA02, and DINOv2) as backbones under the GTA5 $\rightarrow$ \{Cityscapes, BDD-100K, Mapillary\} setting.}
\label{fig:vfms}
\end{figure*}

\begin{table}[t]
    \centering
    \renewcommand\arraystretch{1.5} 
    \setlength\tabcolsep{6pt} 
    \resizebox{1.0\linewidth}{!}{ 
        \begin{tabular}{lcccccc}
            \toprule 
            \multirow{2}{*}{\textbf{Methods}} &
            \multicolumn{4}{c}{\textbf{ACDC}} & \multirow{2}{*}{\textbf{BDD}} & \multirow{2}{*}{\textbf{Map}} \\
            \cline{2-5}
            & \textbf{Fog} & \textbf{Rain} & \textbf{Snow} & \textbf{Night} \\
            \hline
            ISW~\cite{choi2021robustnet} &26.2 &26.7 &22.7  &6.0 &35.4 &38.5\\
            CIN~\cite{learningli}  &29.3  &28.1  &23.7 &12.0 &37.8 &44.4 \\
            \hline
            HGFormer~\cite{ding2023hgformer} & 69.9 & 72.0 & 68.6 & 52.7 & 61.5 & 72.1 \\
            Rein~\cite{wei2024stronger} & 79.5 & 72.5 & 70.6 & 55.9 & 63.5 & 74.0 \\
            \textbf{Ours}  & 80.1 & 74.1 & 71.3 & 56.4 & 65.2 &75.1 \\
            \bottomrule
        \end{tabular}
    }
    \caption{Performance comparison in real-to-real generalization under the GTA5 $\rightarrow$ \{ACDC, BDD-100K, Mapillary\}. \cite{ding2023hgformer}, \cite{wei2024stronger} and ours use the VFM-based backbone (DINOV2). \cite{choi2021robustnet} and \cite{learningli} use the classical-based backbone (Resnet-50~\cite{He2016DeepRL}).}
    \label{tab:real-real}
\end{table}

\subsection{Dataset Description}
Since we introduce SPL for the first time and partition the data into KS and AS, the core of the SPL relies on knowledge transfer by leveraging the capabilities of VFMs. Therefore, the selected dataset can remain the same as in the domain generalization task.
Following the evaluation protocol of recent state-of-the-art DG methods, we conduct extensive experiments on various wide-ranging segmentation datasets. These datasets encompass both synthetic datasets and real-world datasets. 
In order to comprehensively evaluate the generalization capability of our proposed method, we specifically assess its performance on the adverse-condition dataset ACDC~\cite{Sakaridis2021ACDCTA}.

\noindent\textbf{Synthetic Datasets.} We adopt synthetic datasets as knowledge sets.
Among these synthetic datasets, GTA5~\cite{Richter2016PlayingFD} stands out as a significant resource. It consists of a large-scale collection of 24,966 driving-scene images, each with a size of 1914$\times$1052. These images are rendered using the Grand Theft Auto V game engine, allowing for automatic annotation into distinct semantic categories at the pixel level. Another widely used synthetic dataset is SYNTHIA~\cite{Ros2016TheSD}, which comprises 9,400 synthetic images with a size of 1280$\times$760. For our experiments, we specifically adopt the SYNTHIA-RAND-CITYSCAPES subset, following the setting of previous works.

\begin{table}[t]
    \centering
    \renewcommand\arraystretch{1.5} 
    \setlength\tabcolsep{8pt} 
    \resizebox{0.9\linewidth}{!}{ 
        \begin{tabular}{lcccc}
            \toprule
            \textbf{Methods} & \textbf{Citys} & \textbf{BDD} & \textbf{Map} & \textbf{Avg.} \\
            \hline
            RobustNet~\cite{choi2021robustnet}  & 37.7 & 34.1 & 38.5 & 36.8 \\
            PintheMem~\cite{kim2022pin}  & 44.5 & 38.1 & 42.7 & 41.8 \\
            SAN-SAW~\cite{peng2022semantic}  & 42.1 & 37.7 & 42.9 & 40.9 \\
            WildNet~\cite{lee2022wildnet}  & 43.7 & 39.9 & 43.3 & 42.3 \\
            DIGA~\cite{wang2023diga}  & 46.4 & 33.9 & 43.5 & 41.3 \\
            SPC~\cite{huang2023style}  & 46.4 & 43.2 & 48.2 & 45.9 \\
            \hline
            Rein (EVA02)~\cite{wei2024stronger}  & 63.5 & 60.7 & 63.9 & 62.7 \\
            Ours (EVA02)   & \textbf{64.8} & \textbf{62.0} & \textbf{65.1} & \textbf{64.0} \\
            \bottomrule
        \end{tabular}
    }
    \caption{Performance comparison of the VFM-based method with other DGSS methods under the \{GTA5 + Synthia\} $\rightarrow$ \{Cityscapes, BDD-100K, Mapillary\} setting.}
    \label{tab:performance_comparison1}
\end{table}

\noindent\textbf{Real-world Datasets.}
To verify the generalization ability of models, we utilize four real-world datasets: Cityscapes~\cite{Cordts2016TheCD}, BDD-100k~\cite{Yu2020BDD100KAD}, Mapillary~\cite{Neuhold2017TheMV}, and ACDC~\cite{Sakaridis2021ACDCTA} during testing. The validation sets of these datasets are used to evaluate the performance of our method for comparison with other approaches. Cityscapes is a large-scale urban scene dataset. It contains 2,975 images in the training set and 500 images in the validation set with the resolution of 2048$\times$1024. BDD-100k is a real-world dataset that has 7000 training images and 1000 validation images. The images are collected from the US with a resolution of 1280$\times$720. Furthermore, we utilize the Mapillary dataset, which focuses on street view imagery. It consists of 18,000 training images and 2,000 validation images. The images in Mapillary have a resolution of 1920$\times$1080. ACDC is a real-world dataset for semantic scene understanding in adverse conditions, including fog, rain, snow, and night. Each set within the ACDC dataset contains 100 images for validation, except for the night set, which contains 106 images.

\begin{figure*}[t]
  \centering
   \includegraphics[width=1.00\textwidth]{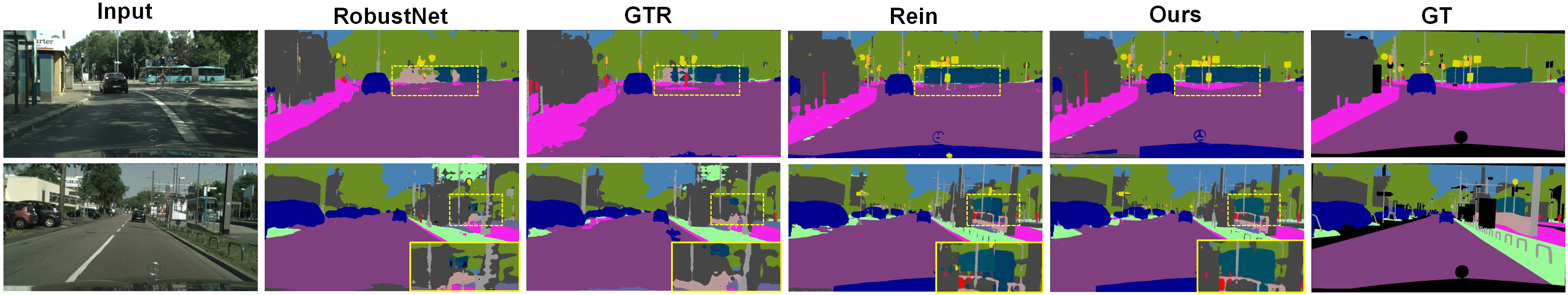}
   \caption{Qualitative results on generalized semantic segmentation. The model is trained on GTA5 and generalized to Cityscapes with DINOV2.}
   \label{result}
\end{figure*}

\begin{figure}[t]
  \centering
\includegraphics[width=1.00\columnwidth]{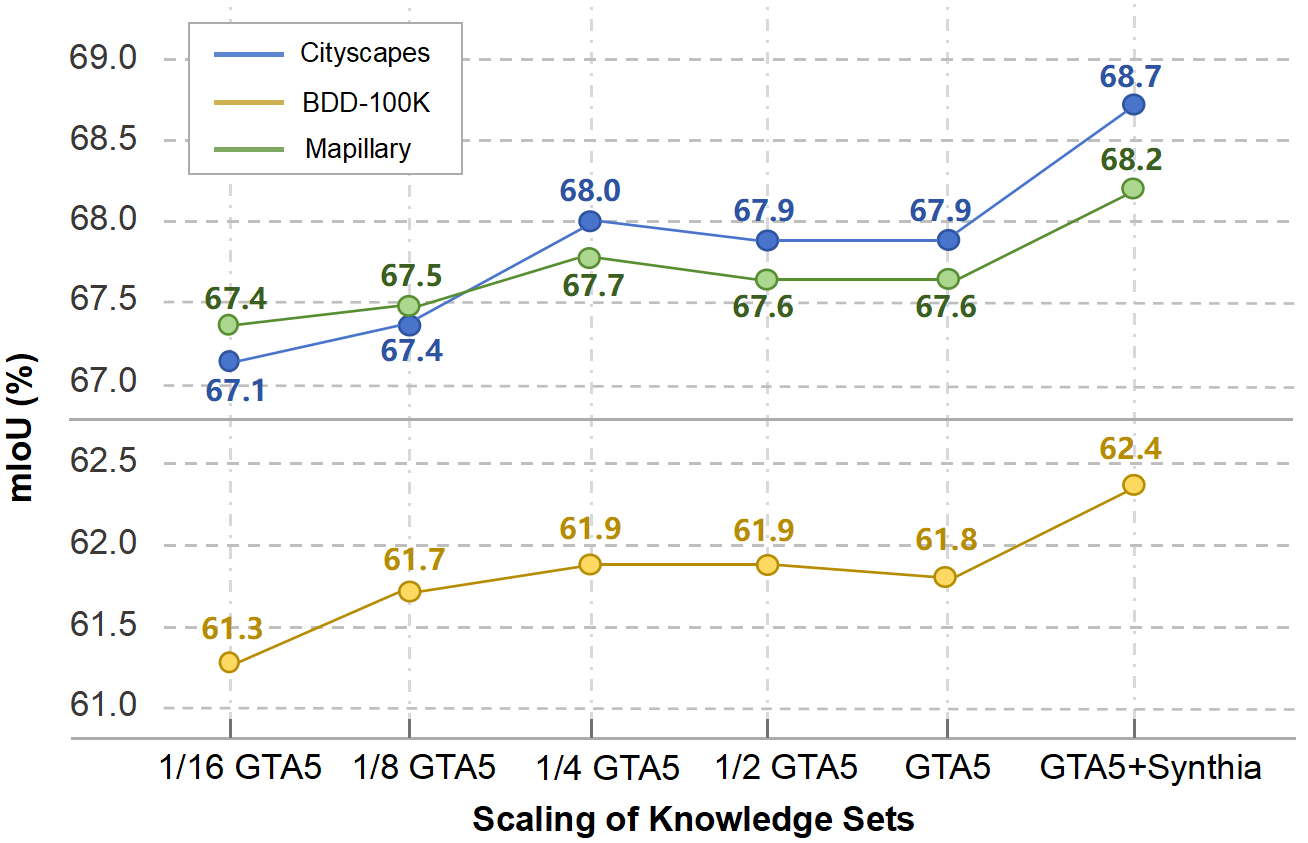}
   \caption{Ablation of different knowledge set scaling on GTA5 $\rightarrow$ \{Cityscapes, BDD-100K, Mapillary\} using DINOV2 backbone.}
   \label{scaleing1}
\end{figure}

\subsection{Implementation Details}
In the SPL, five distinct vision foundation models are used for generalized semantic segmentation: MAE, CLIP, SAM, EVA02, and DINOv2. We implement our approach using the MMSegmentation codebase~\cite{MMSegmentation2020}. For the task head, we incorporate Mask2Former~\cite{Cheng2021MaskedattentionMT}, a widely adopted segmentation decoder, in conjunction with various VFMs as the backbone. We implement the model with a batch size of 4 on a single NVIDIA RTX3090 with 24 GB memory.

\noindent\textbf{Parameter Settings.}
During the training phase, we employ the AdamW optimizer~\cite{kingma2015adam} with a decay of $0.05$. 
Note that in the following experiments, the VFMs employed by our method are kept frozen. The network is trained only on our proposed modules and the decoder, with a learning rate of \(1 \times 10^{-4}\) applied to both.

\noindent\textbf{Data Preparation.}
For an efficient training process, we configure the training for 40K iterations and crop images to a resolution of \(512 \times 512\). Our method utilizes only basic data augmentation techniques, in line with the Mask2Former~\cite{Cheng2021MaskedattentionMT}. 
 
\noindent\textbf{Evaluation Metric.} For the evaluation of segmentation performance, we use the PASCAL VOC Intersection over Union (IoU)~\cite{Everingham2014ThePV} as the evaluation metric. MIoU is the mean value of IoUs across all categories.

\begin{figure}[t]
  \centering
\includegraphics[width=1.00\columnwidth]{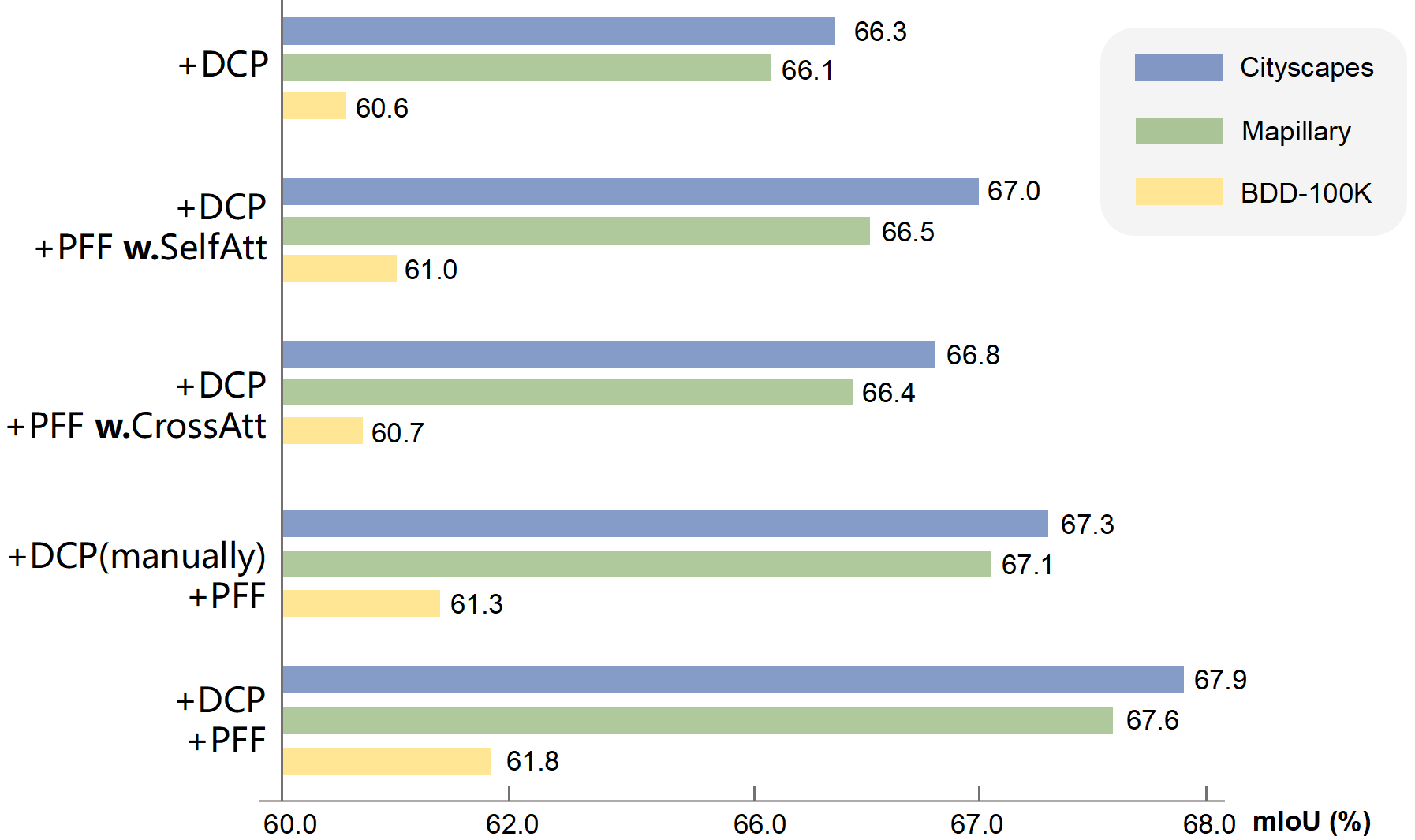}
   \caption{Ablation of key components on GTA5 $\rightarrow$ \{Cityscapes, BDD-100K, Mapillary\} using DINOV2 backbone.}
   \label{tab:ablation}
\end{figure}

\subsection{Comparison with State-of-the-Arts}
\label{sec:comparison}
While we are the first to introduce the concept of Set Pivot Learning, it is noteworthy that recent advancements have incorporated VFMs as foundational elements, particularly for domain generalization semantic segmentation (DGSS). Especially, Rein~\cite{wei2024stronger}, which first assesses and harnesses VFMs in DGSS, achieves impressive performance. Therefore, to validate the effectiveness of our method, we conduct a comparative analysis against the current state-of-the-art. 

\noindent\textbf{SPL in Synthetic-to-Real.}
Table~\ref{tab:syn-real} presents the comparative experiments conducted under the benchmarking. These methods are trained on synthetic data, and then generalized to real-world datasets. 
In particular, the VFM-based approach, Rein, has comprehensively surpassed the previous DGSS approaches. In the Fig.~ \ref{result}, we present the qualitative semantic segmentation results.
In addition, as shown in Fig.~\ref{fig:vfms}, we further incorporate additional VFMs as backbone networks, with the baseline representing the results of the models in their frozen state. The experimental results demonstrate that our method significantly enhances the generalization performance of models from synthetic environments to real-world scenarios.

\begin{figure*}[t]
  \centering
   \includegraphics[width=1.00\textwidth]{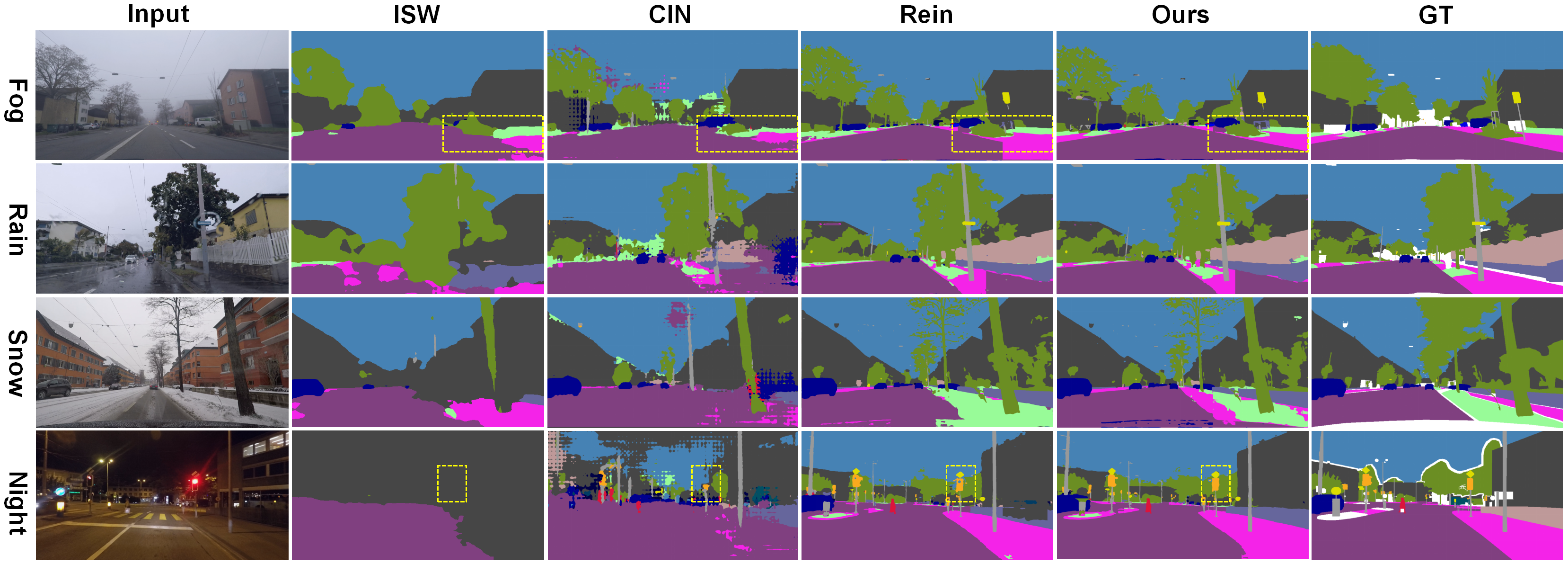}
   \caption{Qualitative results on generalized semantic segmentation under adverse weather. The model is trained on GTA5 and generalized to ACDC with DINOV2.}
   \label{acdcresult}
\end{figure*}

\noindent\textbf{SPL in Real-to-Real.}
Real-to-real generalization, where a model trained on one real-world dataset is tested on others, is essential for practical deployment in diverse environments.  To evaluate the effectiveness of our method in real-to-real, we conduct experiments under normal weather and adverse weather settings: Cityscapes $\rightarrow$ \{BDD-100K, Mapillary\} and Cityscapes $\rightarrow$ \{ACDC-fog, ACDC-rain, ACDC-snow, ACDC-night\}. The comparative experiments, presented in Tab.~\ref{tab:real-real}, demonstrate that our method achieves superior generalization across a range of real-world datasets.
Fig.~\ref{result} and Fig.~\ref{acdcresult} provide the qualitative semantic segmentation results under normal and adverse weather settings. Specifically, compared to other methods, our approach is more accurate in the segmentation of objects, which are circled in yellow.

\begin{figure*}[t]
  \centering
   \includegraphics[width=1.00\textwidth]{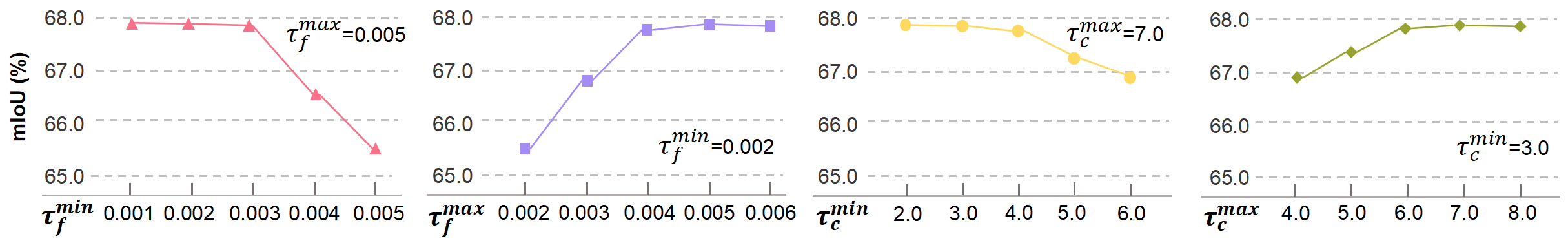}
   \caption{Ablation of the parameter settings in the DCP module with GTA5 $\rightarrow$ Cityscapes using DINOV2 backbone.}
   \label{parresult}
\end{figure*}

\subsection{Multiple Knowledge Sets generalization}
As aforementioned in our definition of knowledge set (KS), all available training data are considered part of the KS, which is no longer confined to a specific domain where data share the same distribution. 
To further validate the effectiveness of our method in the context of multiple sets, we evaluate its performance against other DGSS methods in the \{GTA5 + Synthia\} $\rightarrow$ \{Cityscapes, BDD-100K, Mapillary\} setup, where models are fine-tuned on GTA5 and Synthia and then evaluated on Cityscapes, BDD-100K, and Mapillary. 
Tab.~\ref{tab:performance_comparison1} presents the results for using EVA02~\cite{fang2023eva02}, demonstrating that our method yields a significant improvement in the scenario of involving multiple knowledge sets compared to previous DGSS approaches. Specifically, our proposed method outperforms existing approaches by over 1.2\% in average mIoU across the three application sets, highlighting its the efficacy and superiority in the context of multiple knowledge sets.

\subsection{Ablation Study}
We conduct extensive ablation studies to evaluate the various components of the proposed method in the setting: GTA5~$\rightarrow$ \{Cityscapes, BDD-100K, Mapillary\}.

\noindent\textbf{Effect of Knowledge Set Scaling.}
Building on the foundation of large-scale pretrained models, we further investigate the impact of the knowledge set (i.e., training dataset) size on the robustness of models in downstream tasks. Fig.~\ref{scaleing1} illustrates that the model achieves optimal mIoU performance when using a quarter of GTA5 dataset, while maintaining comparable results across other dataset sizes. By leveraging prompt-based fine-tuning to steer the generalization capacity of VFMs toward specific downstream tasks, our approach reduces sensitivity to training data scaling. We further conduct ablation studies across multiple datasets, including SYNTHIA, to evaluate model performance.

\begin{table}[t]
    \centering
    \renewcommand\arraystretch{1.3} 
    \setlength\tabcolsep{6pt}
    \resizebox{0.48\textwidth}{!}{ 
        \begin{tabular}{l|ccccccc}
            \toprule
            \textbf{Setting of $N$} & 10 & 20 & 30& 40& 50& 60 & 70\\
            \hline
            Citys (mIoU)    & 67.2& 67.8& 67.9& 67.7 & 67.1 & 66.8 & 66.2\\
            \bottomrule
        \end{tabular}
    }
    \caption{Ablation of the maximum output classes $N$ in the DCP module with DINOV2 backbone.}
    \label{tab:numberaltation}
\end{table}

\noindent\textbf{Analysis of DCP.}
We evaluate of the effectiveness of DCP, as illustrated in Fig.~\ref{tab:ablation}. In the experiment, we set $\tau_{f}^{\min} = 0.002$, $\tau_{f}^{\max} = 0.005$, and $\Delta \tau_f = 0.001$, while $\tau_{c}^{\min} = 3.0$, $\tau_{c}^{\max} = 7.0$, and $\Delta \tau_c = 0.5$. 
We further evaluate the robustness of our method with respect to these parameters in Fig.~\ref{parresult}, demonstrating its consistent performance across a range of values.
For example, when $\tau_{f}^{\max} = 0.005$, the method achieves comparable performance across the range from $\tau_{f}^{\min} = 0.001$ to $0.003$.
However, when $\tau_{f}^{\min} = 0.004$ or $0.005$, the number of prompts decreases sharply, leading to a corresponding drop in mIoU performance.
Notably, these parameters can be flexibly adjusted, making the approach suitable for a variety of downstream tasks.
In particular, for the maximum output classes $N$, we have obtained comparable experimental results from 20 to 40 as shown in Tab.~\ref{tab:numberaltation}, which also proves the robustness of our method.
In addition, for the DCP module, we compare the performance of manually inputted prompts with our automatically generated prompts. Specifically, we manually provide prompts based on the categories that appear in the image, such as ``car, road, pedestrian, sidewalk...”. 
In Fig.~\ref{tab:ablation}, the experimental result (manually) shows only a slight decrease in mIoU, which nonetheless underscores the effectiveness of importing class prompts. However, the manual input of prompts is impractical in real-world applications, as it is highly time-consuming and labor-intensive when dealing with large-scale datasets.

\noindent\textbf{Analysis of PFF.}
As shown in Fig.~\ref{tab:ablation}, the terms “PFF w.SelfAtt” and “PFF w.CrossAtt” refer to configurations where only the text-to-text self-attention or image-to-text cross-attention modules is used for the PFF, respectively. 
The ablation studies demonstrate that our designed Cross-Modal Interaction module significantly enhances performance in the DGSS. Within the PFF module, we further investigate the impact of the learnable token length on the module's performance. As illustrated in Tab.~\ref{tab:tokenablation}, our proposed model achieves optimal performance at a token length of 75. Notably, the results reveal that token lengths in the range of 50 to 100 yield comparable performance, suggesting a relative insensitivity of the module to this parameter. Based on these findings, we adopt a token length of 75 as the optimal configuration for our implementation.

\begin{table}[t]
    \centering
    \renewcommand\arraystretch{1.5} 
    \setlength\tabcolsep{9pt} 
    \resizebox{1.0\linewidth}{!}{ 
        \begin{tabular}{lcccccc}
            \toprule 
            \multirow{2}{*}{\textbf{Datasets}} &
            \multicolumn{6}{c}{\textbf{Length of token}}\\
            \cline{2-7}
            & 25 & 50 & 75 & 100 & 125 & 150 \\
            \hline
            Citys &67.2 &67.8 &67.9  &67.7 &67.4 &67.3 \\
            BDD  &61.3  &61.8  &61.8 &61.6 &61.2 &61.1  \\
            Map  &66.9  &67.5  &67.6 &67.5 &67.1 &67.2 \\
            \hline
            Avg. (mIoU)   & 65.1& 65.7& 65.8& 65.6 & 65.2 & 65.2\\
            
            \bottomrule
        \end{tabular}
    }
    \caption{Ablation of the length of learnable tokens in the PFF module with DINOV2 backbone.}
    \label{tab:tokenablation}
\end{table}

\section{Conclusion}
In this work, we introduced Set Pivot Learning, redefining the domain generalization in downstream tasks using VFMs. 
SPL aims to hone task-specific representations while using VFM-based knowledge as the pivot through dynamic adaptation.
By redefining DG challenges in the context of large-scale pretrained data, we leverage VFMs to address critical issues in robustness and adaptability. 
This process shifts the focus from traditional feature transfer to feature refinement and task-specific adaptation. We proposed a dynamic prompt fine-tuning method, which combines a Dynamic Class-aware Prompter to extract scene-specific prompts with a Prompt-guided Feature Focuser that employs multimodal attention to refine VFM features. 
Through extensive experiments and ablation studies on benchmark datasets, our method demonstrates clear performance gains over existing methods, particularly in the field of generalized segmentation.

\small{
\bibliographystyle{IEEEtran}
\bibliography{IEEEabrv, main
}
}

\vfill
\end{document}